\documentclass[final]{cvpr}

\usepackage{times}
\usepackage{epsfig}
\usepackage{graphicx}
\usepackage{amsmath}
\usepackage{amssymb}
\usepackage{multirow}
\usepackage{algorithm}
\usepackage{algorithmic}
\usepackage{subfigure}
\usepackage{comment}
\usepackage{diagbox}
\usepackage{makecell}
\usepackage[pagebackref=true,breaklinks=true,colorlinks,bookmarks=false]{hyperref}

\def\cvprPaperID{4621}
\def\confYear{CVPR 2021}

\begin{document}

\title{Beyond Max-Margin: Class Margin Equilibrium for Few-shot Object Detection}

\author{Bohao Li$^{1}$\thanks{Equal Contribution.} \and Boyu Yang$^{1}$\footnotemark[1] \and Chang Liu$^1$ \and Feng Liu$^1$ \and Rongrong Ji$^{2,3,4}$ \and Qixiang Ye$^{1,4}$\thanks{Corresponding Author.}
\and PriSDL, EECE, University of Chinese Academy of Sciences, 100049, China.$^1$  \and  MAC, Department of Artificial Intelligence, School of Informatics, Xiamen University, 361005, China.$^2$ \and Institute of Artificial Intelligence, Xiamen University, 361005, China.$^3$ \and Peng Cheng Laboratory, Shenzhen, China.$^4$

\and\tt\small \{libohao20, yangboyu18, liuchang615, liufeng20\}@mails.ucas.ac.cn 
\and\tt\small rrji@xmu.edu.cn \and\tt\small qxye@ucas.ac.cn
}

\maketitle

\begin{abstract}
Few-shot object detection has made substantial progress by representing novel class objects using the feature representation learned upon a set of base class objects.
However, an implicit contradiction between novel class classification and representation is unfortunately ignored.
On the one hand, to achieve accurate novel class classification, the distributions of either two base classes must be far away from each other (max-margin).
On the other hand, to precisely represent novel classes, the distributions of base classes should be close to each other to reduce the intra-class distance of novel classes (min-margin).
%
In this paper, we propose a class margin equilibrium (CME) approach, with the aim to optimize both feature space partition and novel class reconstruction in a systematic way.
CME first converts the few-shot detection problem to the few-shot classification problem by using a fully connected layer to decouple localization features.
%
CME then reserves adequate margin space for novel classes by introducing simple-yet-effective class margin loss during feature learning.
Finally, CME pursues margin equilibrium by disturbing the features of novel class instances in an adversarial min-max fashion.
Experiments on Pascal VOC and MS-COCO datasets show that CME significantly improves upon two baseline detectors (up to $3\sim5\%$ in average), achieving state-of-the-art performance. Code is available at \href{https://github.com/Bohao-Lee/CME}{\color{magenta}https://github.com/Bohao-Lee/CME}.

\end{abstract}

\section{Introduction}

In the past few years, we witnessed the great progress of visual object detection~\cite{YOLO,YOLOV2,FRCNN,FasterRCNN}. This is attributed to the availability of large-scale datasets with precise annotations and convolutional neural networks (CNNs) capable of absorbing the annotation information. However, annotating a large amount of objects is expensive and laborious~\cite{CMIL2019,MinEntropy2019,FreeAnchor2019,zhu2017soft}. It is also not consistent with cognitive learning, which can build a precise model using few-shot supervisions~\cite{CompositionalRepre-ICCV2019}.  

\begin{figure}[t]
\centering
\includegraphics[width=1\linewidth]{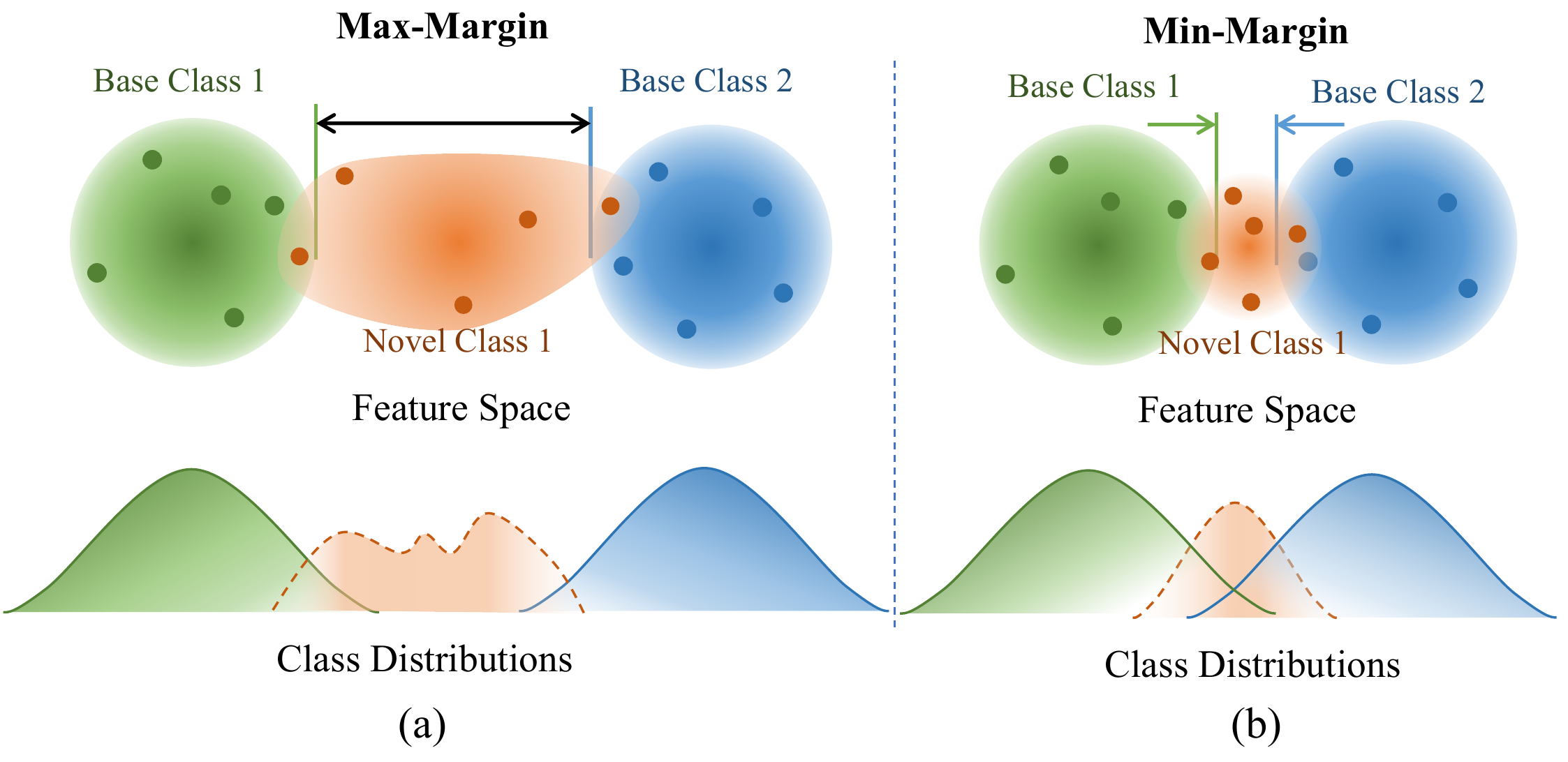}
\caption{The contradiction of representation and classification in few-shot object detection. (a) To separate the classes with each other, either two base classes requires to be far away from each other (max-margin), which aggregates the intra-class distance of novel classes. (b) To precisely represent novel classes, the distributions of base classes should be close to those of novel classes (min-margin), which improves the difficulty of classification.}
\label{fig:motivation}
\end{figure}

Few-shot detection, which simulates the way that human learns, has attracted increasing attention. Given base classes of sufficient training data and novel classes of few supervisions, few-shot detection trains a model to simultaneously detect objects from both base and novel classes. To this end, a majority of works divided the training procedure to two stages: base class training (representation learning) and novel class reconstruction (meta training). In representation learning, the sufficient training data of base classes are used to train a network and constructs a representative feature space. In meta training, the network is finetuned so that the novel class objects can be represented within the feature space. Among the earliest work, Kang \etal~\cite{FeatureReweighting} proposed applying channel-attended feature re-weighting for semantic enforcement. In the two-stage framework, Wang \etal~\cite{MetaDet} and Yan \etal~\cite{MetaRCNN} contributed early few-shot detection methods. Wang \etal~\cite{Frustratingly} and Wu \etal~\cite{MPSR} proposed freezing the backbone network and reconstructing the novel classes using the classifier weights during detector finetuning. 

Despite the substantial progress, the implicit contradiction between representation and classification is unfortunately ignored. To separate the classes, the distributions of two base classes requires to be far away from each other (max-margin), which however aggregates the diversity of novel classes, Fig.~\ref{fig:motivation}(a). To precisely represent novel classes, the distributions of base classes should be close to each other (min-margin), which causes the difficult of classification, Fig.~\ref{fig:motivation}(b). How to simultaneously optimize novel class representation and classification in the same feature space remains to be elaborated.

In this paper, we propose a class margin equilibrium (CME) approach, with the aim to optimize feature space partition for few-shot object detection with adversarial class margin regularization. For the object detection task, CME first introduces a fully connected layer to decouple localization features which could mislead class margins in the feature space. CME then pursues a margin equilibrium to comprise representation learning and feature reconstruction. Specifically, during base training, CME constructs a feature space where the margins between novel classes are maximized by introducing class margin loss. During network finetuning, CME introduces a feature disturbance module by truncating gradient maps. With multiple training iterations, class margins are regularized in an adversarial min-max fashion towards margin equilibrium, which facilities both feature reconstruction and object classification in the same feature space.

 The contributions of this study include:
\begin{itemize}
    \item We unveil the representation-classification constriction hidden in few-shot object detection, and propose a feasible way to alleviate the constriction from the perspective of class margin equilibrium (CME).
    
    \item We design the max-margin loss and feature disturbance module to implement class margin equilibrium in an adversarial min-max fashion.
    
    \item We convert the few-shot detection problem to a few-shot classification problem by filtering out localization features, We improve the state-of-the-art with significant margins upon both one-stage and two-stage baseline detectors.
\end{itemize}

\begin{figure*}[t]
\centering
\includegraphics[width=1\linewidth]{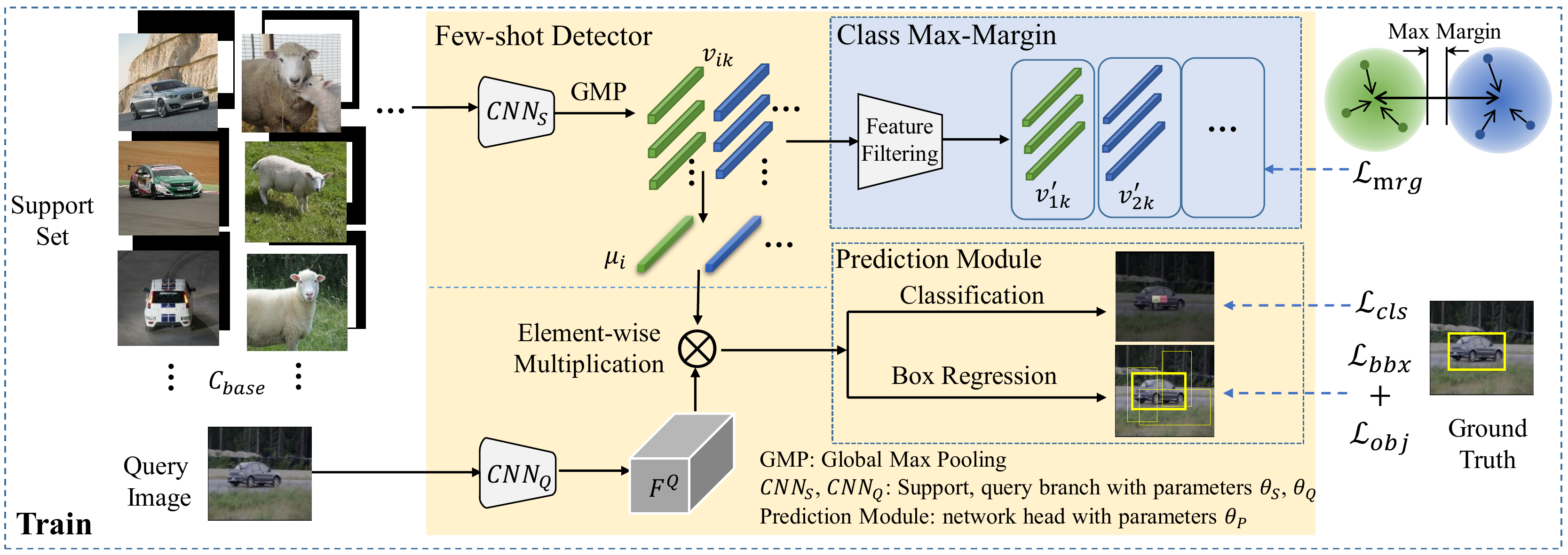}
\caption{Framework of the proposed few-shot detection which consists of a support branch and a query branch. This figure only illustrates base class training driven by detection loss and max-margin loss.}
\label{fig:flowchart}
\end{figure*}

\section{Related Works}
\subsection{Object Detection}
CNN-based methods have significantly improved the performance of object detection. While the one-stage methods, $e.g.$, YOLO~\cite{YOLO,YOLOV2} and SSD~\cite{SSD}, have higher detection efficiency, the two-stage methods, $e.g.$, Faster R-CNN~\cite{FRCNN,FasterRCNN} and FPN~\cite{FPN} report higher performance, usually. Relevant CNN-based detectors provided fundamental techniques, $e.g.$, RoI pooling~\cite{FRCNN} and multi-scale feature aggregation~\cite{FPN}, which benefit few-shot object detection. However these methods generally require large amounts of training data, which hinders their applications in practical scenarios.

\subsection{Few-shot Learning}
Existing few-shot learning methods can be broadly categorized as either: metric learning~\cite{MatchNetwork16,Compare2018,CollectSelect19,DeepEMD,PMMs,Liu_2021_CVPR,Harmonic,AMN}, meta-learning ~\cite{LearningToLearn16,Optimization17,MAML17,TaskAgnosticMeta19}, or data augmentation~\cite{Hallucinating17,Imaginary18}.
Metric learning methods train networks to predict whether two images/regions belong to the same category. Meta-learning approaches specify optimization or loss functions which force faster adaptation of parameters to new categories with few examples. The data argumentation methods learn to generate additional examples for unseen categories. In the metric learning framework, prototypical models converted the spatial semantic information of objects to convolutional channels. 
%
In existing studies, it was observed that class margin has a great impact to classifiers when required to guarantee model discriminability under few supervisions. Li~\etal.~\cite{AdaptiveMargin} proposed adaptive margin loss to improve model generalization ability. They further developed a class-relevant additive margin loss considering the semantic similarity between image pairs.
%
However, solely pursuing max-margin could be infeasible because the novel classes required to be reconstructed with the base classes and large margin would improve the diversity of novel class samples.
%
Liu~\etal. ~\cite{NegativeMargin} introduced negative class margin to benefit representation of novel classes. Existing studies inspire us to re-think the max-margin principle in few-shot settings, to comprise discriminability and representation capability of features.

\subsection{Few-shot Object Detection}
Following the meta learning methods, Kang \etal~\cite{FeatureReweighting} contributed an early few-shot detection method, which fully exploited training data from base classes while quickly adapting the detection prediction network to predict novel classes. Yan~\etal~\cite{MetaRCNN} proposed meta-learning over RoIs, enabling Faster R-CNN be a meta-learner for few-shot detection. 
%
Wu~\etal.~\cite{MPSR} proposed positive sample refinement to enrich object scales for few-shot detection. Despite of the progress, the discriminability and representation equilibrium between novel and base classes remain unsolved. Furthermore, most existing methods treat few-shot detection as a few-shot classification problem, ignoring the role of features for object localization. 

\section{The Proposed Approach}

\subsection{Few-shot Detection Framework}
\textbf{Problem Definition.}
Given base classes $C_{base}$ of sufficient training data and novel classes $C_{novel}$ of few supervisions, few-shot detection aims to train a model that can simultaneously detect objects from both base and novel classes. As shown in Fig.\ \ref{fig:flowchart}, a detection network is first trained with $C_{base}$ to construct feature representation. The network is then finetuned with both $C_{base}$ and $C_{novel}$ to represent the few-shot instances from novel classes. In what follows, we describe the proposed method by using meta YOLO~\cite{FeatureReweighting} as the baseline detector. Our approach can be applied to two-stage few-shot detectors~\cite{MPSR} in a plug-and-play fashion.

For both detector training and finetuning, the dataset $\mathcal{D}$ (either base classes or novel classes) is divided to a support set $\mathcal{S}$ and a query set $\mathcal{Q}$. $\mathcal{D} = \mathcal{S}\cup\mathcal{Q} = \{I^S, M^S\}\cup \{I^Q, M^Q\}$, where $I^S$ denotes support images with a mask annotations $M^S$, which are generated according to object bounding-boxes. $I^Q$ denotes the query images with ground-truth bounding boxes $M^Q$. Given $N$ classes, each of which has $K$ annotated instances,  $\mathcal{S}$ can be further denoted as $\{\{I_{ik}^S,M_{ik}^S\},i=1,...,N, k=1,...,K\}$, where $i$ indexes the class and $k$ indexes instance, and $I_{ik}^S\in \mathbb{R}^{W\times H \times 3}$.

The network consists of a support branch (Fig.~\ref{fig:flowchart} (upper)) a query branch (Fig.~\ref{fig:flowchart} (lower)). On the support branch, the support images $I^S$ and their bounding-boxes $M^S$ are fed to the CNN to extract convolutional feature maps. With a global max pooling (GMP) operation, the feature maps are squeezed to prototype vectors $v_{ik} = f_{\theta_S}(I^S\oplus M^S)$, where $f_{\theta_S}(\cdot)$ denotes the network of the support branch with parameters $\theta_S$ and $\oplus$ the concatenate operation. The mean prototypes for the $i$-th class is calculated by $\mu_i =\frac{1}{K} \sum_{k=1}^{K} v_{ik}$, indicating the semantics of the object class. 
On the query branch, convolutional features $F^Q= f_{\theta_Q}(I^Q)$ are extracted for the query images $I^Q$, where $\theta_Q$ denotes the query branch network parameters. The features are activated by multiplying with the prototype vectors $\{\mu_{i}\}$ through a pixel-wise multiple operation. 
The activated features are fed to a prediction (classification and box regression) module and output $\mathcal{P}_{\theta_P}(F^Q\otimes \mu_i)$, where $\theta_P$ denotes the prediction module parameters, $\otimes$ means element-wise multiplication. For the general object detection task, the target of the prediction results are expected to match the ground-truth bounding box area $M^Q$, through minimizing the following loss
\begin{align}
   \mathop{\arg\min}_{\theta} \mathcal{L}_{det}(\mathcal{P}_{\theta_P}(F^Q\otimes u_i),M^Q) \label{eq:det_loss},
\end{align}
where $\theta=\theta_S \cup \theta_Q \cup \theta_P$. $\mathcal{L}_{det}$ is the object detection loss, defined as $\mathcal{L}_{det} = \mathcal{L}_{cls}+\mathcal{L}_{bbx}+\mathcal{L}_{obj}$, where $\mathcal{L}_{cls}$, $\mathcal{L}_{reg}$, and $\mathcal{L}_{obj}$ respectively denote the classification, regression and anchor confidence loss~\cite{FeatureReweighting,YOLOV2}.

\textbf{Feature Filtering.} For object detection, the convolutional features incorporate both localization features and classification features. While the classification features are class dependent, the localization features are independent to object classes, and therefore tend to perturb class margins. To filtering out the localization features, a fully connected layer is used to decouple localization features, as $v_{ik}^{'} = \mathcal{FC}(v_{ik})$, to convert the few-shot detection problem to a pure few-shot classification problem, Fig.\ \ref{fig:flowchart}. Driven by the max-margin loss, the localization features are filtered out during detector training.

\subsection{Base Training: Class Max-margin}

\textbf{Max-Margin Loss.}
In the base training stage, the sufficient data of base classes are used to train the network and construct a representative feature space. As the object detection network is a discriminative model, the whole feature space will be divided into multiple sub-spaces each of which is occupied with a class. 
In the finetuning stage, novel classes will be embedded to the feature space, often to the margin space between base classes. To avoid aliasing, the margin space between base classes should be big enough to accommodate novel classes, Fig.\ \ref{fig:motivation}(a), $i.e.$, class max-margin. 

To pursue class max-margin, prototype vectors of the base classes require to be close to their mean prototypes ($i.e.$, minimum intra-class variance) while those of different classes be far away from each other ($i.e.$, maximum inter-class distance). Given the prototype vector $v_{ik}^{'}$ for the $k$-th instance, the mean prototype vector of the $i$-th class is calculated as $\mu_i^{'} =\frac{1}{K} \sum_{j=0}^{k-1} v_{ij}^{'}$, which represents the semantics of the class. The intra-class distance is $D_{i}^{Intra} =\sum_{j=0}^{K-1} ||v_{ij}^{'}-\mu_i^{'}||_2^2$. The inter-class margin distance is calculated as  $D_{i}^{Inter} =\min_{j,j\neq i}||\mu_j^{'}-\mu_i^{'}||_2^2$. Generally speaking, margin is defined as the distance between the decision boundary and the sample of shortest distance to the boundary. For the feature space constructed by CNN, it is hard to directly calculate the margin $\mathcal{M}_{i,i'}$ between two classes. As an approximation, we first calculate the upper and lower bounds of $\mathcal{M}_{i,i'}$ and have
\begin{align}
D_i^{Inter}-D_i^{Intra}-D_{i'}^{Intra}\le \mathcal{M}_{i,{i'}} \le D_i^{Inter},
\label{eq:bound}
\end{align}
which indicates that the upper bound of margin is the inter-class distance while the lower bound is the inter-class distance subtracting intra-class distance. According to Eq.~\ref{eq:bound}, max-margin can be approximated by maximizing the upper and lower bounds of margins, as
\begin{align}
   \arg\max_\theta \mathcal{M}_{i,i'} \simeq\arg\max_\theta  \mathcal{L}_{mrg} = \frac{\sum_i^{N} D_{i}^{Intra}}{\sum_i^{N} D_{i}^{Inter}},
   \label{eq:margin_loss}
\end{align}
where $N$ denotes the class number.

\textbf{Detector Training.}
In base training, a support set and a query set are constructed for base classes by randomly selecting training samples, $\mathcal{S}\cup\mathcal{Q} \subseteq  \mathcal{D}_{C_{base}}$. The detection network is trained by optimizing both the object detection loss and the max-margin loss, as
\begin{align}
   \arg\min_{\theta} \mathcal{L}_{trn} = \mathcal{L}_{det} + \lambda \mathcal{L}_{mrg} \label{eq:loss_train},
\end{align}
where $\lambda=1.0$ is an experimentally determined regularization factor to balance the two loss functions.

\begin{figure}[t]
\centering
\includegraphics[width=1\linewidth]{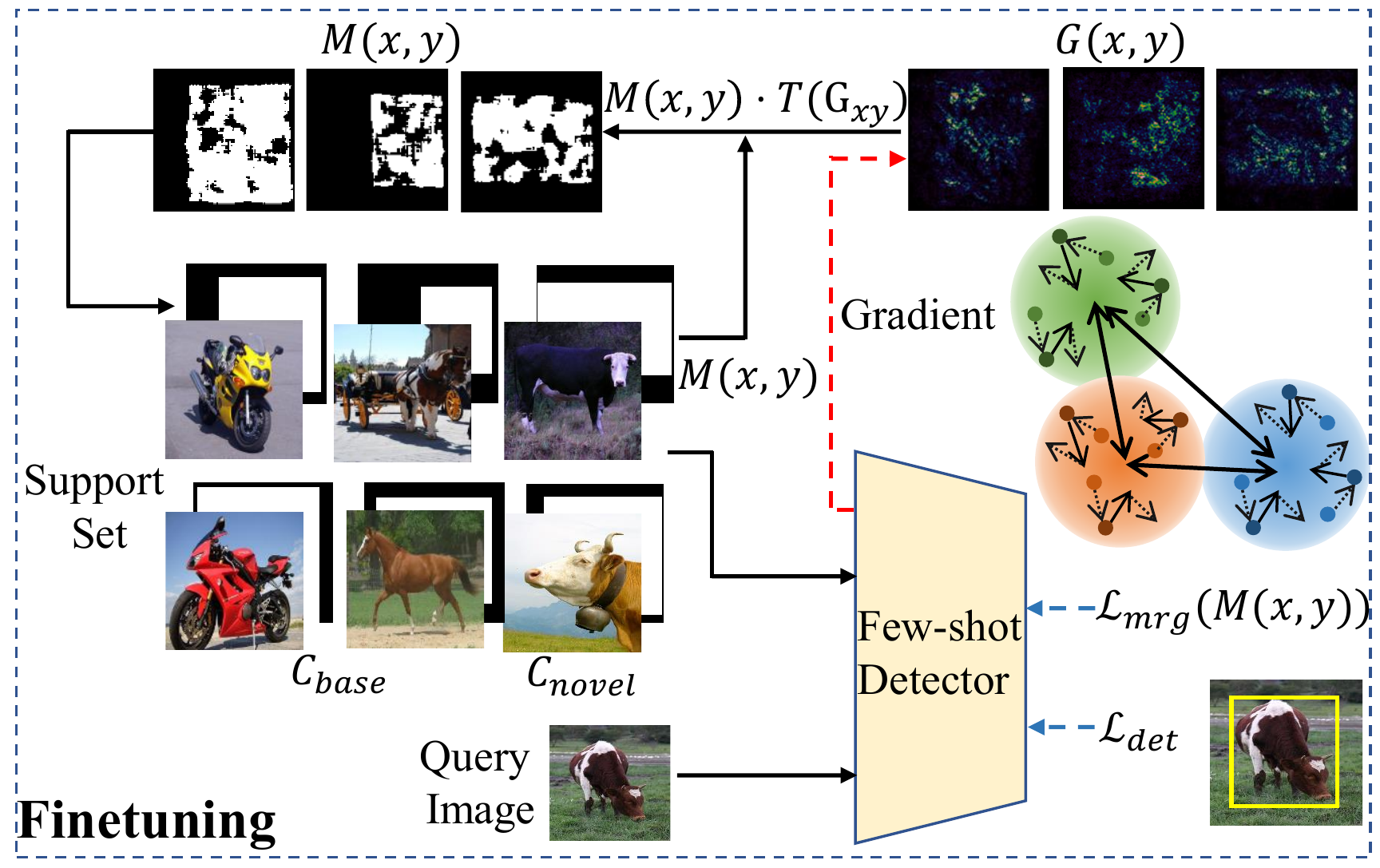}
\caption{Network finetuning with feature disturbance. Feature disturbance is implemented by truncating the gradient maps and re-sampling the training images.}
\label{fig:flowchart_finetune}
\end{figure}

\subsection{Finetuning: Margin Equilibrium}
Finetuning refers to a meta learning procedure, which uses few-shot novel class data to update network parameters. However, without sufficient data, the novel classes cannot significantly change the feature representation, so that novel classes required to be represented by the features learned upon base classes~\cite{NegativeMargin}.  According to Eq.\ \ref{eq:margin_loss}, the margins between base classes in the feature space are required to be large, which improves the $D^{Intra}$ and preserve sufficient margin space for novel classes. However, when the margin between base classes is large, the samples/features from novel classes can be of large diversity, Fig.\ \ref{fig:motivation}(b), which aggregates the difficulty to train the detection model for novel classes. To solve this problem, we propose the margin equilibrium strategy based on feature disturbance. 

\textbf{Feature Disturbance.}
Feature disturbance defines an online data augmentation procedure according to the gradient maps of samples. During finetuning, images of base and novel classes are simultaneously fed to the network detector training driven by the margin loss and detection loss, Fig.\ \ref{fig:flowchart_finetune}. During the back-propagation procedure, the gradient map of a support image is calculated by $G(x,y) = ||\frac{\partial {\mathcal{L}}_{ftn}}{\partial I(x,y)^S}||$, where $G(x,y)\in \mathbb{R}^{W\times H}$ and $||\cdot||$ denotes the norm operation. $\mathcal{L}_{ftn}$ denotes the finetuning loss defined on the detection loss and margin loss. $(x,y)$ is the pixel location. According to the characteristics of CNN, the pixels of larger gradients correspond object parts of larger discrimination ability and contribute more to reduce the finetuning loss. During detector training, the disturbance procedure is carried out to truncate the pixels of large gradient and disturb the finetuned features. This is implemented by re-sampling the ground-truth mask according to the gradient map, as 
\begin{align}
T(G{(x,y)})=\begin{cases}
0& G{(x,y)}\geq \tau(G{(x,y)})\\
1& otherwise
\end{cases},
\end{align}
where $\tau$ is a threshold which controls the ratio of feature disturbance. In experiments, $\tau$ is set to be a dynamic threshold so that top-15\% pixels of large gradient are set to 0. For feature disturbance, the support mask $M^S$ are updated according to the gradients map, as
\begin{align}
M^{S}(x,y) \leftarrow M^{S}(x,y) \cdot T(G(x,y)).
\label{eq:mask_iter}
\end{align}

\begin{algorithm}[t]
\label{alg:onlineEM}
\caption{Detector training and finetuning with CME}
\renewcommand{\algorithmicrequire}{\textbf{Input:}}  
\renewcommand{\algorithmicensure}{\textbf{Output:}}  
\begin{algorithmic}
\REQUIRE ~~\\ 
Support set $\mathcal{S}=\{I^S, M^S\}$, query set $\mathcal{Q}=\{I^Q, M^Q\}$;
\ENSURE ~~\\ 
Network parameters $\theta = \theta_S \cup \theta_Q \cup \theta_P$;

\textbf{Training:}\\
\FOR{($I^S, M^S, I^Q, M^Q$ in $\mathcal{D}_{C_{base}}$)}
\STATE \textbf{Predict} detections and calculate detection loss $\mathcal{L}_{det}$ by Eq.\ \ref{eq:det_loss};\
\STATE \textbf{Calculate} margin loss $\mathcal{L}_{mrg}$ by Eq.\ \ref{eq:margin_loss}\
\STATE \textbf{Update} $\theta$ to minimize the training loss $\mathcal{L}_{trn}$ by Eq.\ \ref{eq:loss_finetune};
\ENDFOR

\textbf{Finetuning:}\\
\FOR{(each $I^S, M^S, I^Q, M^Q$ in $\mathcal{D}_{C_{base}} \cup {D}_{C_{novel}}$)}
\FOR{(iteration}
\STATE \textbf{Predict} detections and calculate the detection loss $\mathcal{L}_{det}$ by Eq.~\ref{eq:det_loss};\
\STATE \textbf{Calculate} margin loss $\mathcal{L}_{mrg}$ by Eq.\ \ref{eq:margin_loss}\
\STATE \textbf{Update} $\theta$ to minimize the finetuning loss $\mathcal{L}_{trn}$ by Eq.\ \ref{eq:loss_finetune} with back-propagation;
\STATE \textbf{Update} support mask $M^S$ according to Eq.\ \ref{eq:mask_iter}.\
\ENDFOR
\ENDFOR

\end{algorithmic}
\end{algorithm}

\textbf{Margin Equilibrium.}
With the above defined feature disturbance strategy, the novel classes $C_{novel}$ are combined with the base classes $C_{base}$ for network parameter finetuning.  Given a batch of support and query images, the network parameters are updated for a few iterations. The iteration number relies on the number ($K$) of training images in each novel class. In each finetuning iteration, the support mask is re-sampled, and the prototype vector is calculated by $v_{ik} = f_{\theta_S}(I^S\oplus M^{S})$ guided by the re-sampled mask $M^{S}$. Accordingly, detector finetuning is performed by minimizing the loss function
\begin{align}
   \mathcal{L}_{ftn} = \mathcal{L}_{det} + \lambda \mathcal{L}_{mrg}(M^S(x,y)).
   \label{eq:loss_finetune}
\end{align}
Meanwhile, according to Eq.~\ref{eq:mask_iter}, the features are disturbed so that prototype vectors within the feature space are re-sampled to occupy the class margin. 

During back-propagation, network parameters are updated to maximize the class margins $\mathcal{M}_{i,i'}$ by optimizing Eq.\ \ref{eq:margin_loss}. In the procedure, $D_i^{Inter}$ increases while $D_i^{Intra}$ decreases. During forward propagation, the support mask $M^S$ is updated according to Eq.\ \ref{eq:mask_iter} and the support image is re-sampled. In this way, the discriminative pixels on the image/features are erased so that the discrimination power of prototype vectors $v_{ik}$ generated by the re-sampled features decreases. As a result, the upper bound $D^{Inter}$ decreases and so does the margin $\mathcal{M}_{i,i'}$.
This actually defines an adversarial learning procedure for min-max margin, as
\begin{align}
\begin{cases}
\arg\underset{\theta}{\max} \mathcal{M}_{i,i'}, &\text{Back Propagation}\\
\arg\underset{M^S(x,y)}{\min} \mathcal{M}_{i,i'}, &\text{Forward Propagation}
\end{cases},
\label{eq:equilibrium}
\end{align}
which pursues class margin equilibrium for base classes and embedded novel classes, detailed in Algorithm {\color{red}1}.

\setlength{\tabcolsep}{4pt}
    \begin{table}[t]
    \begin{center}
    \caption{Ablation study of CME modules for few-shot object detection on Pascal VOC novel classes (split-1). ``MM" denotes max-margin, ``FF" feature filtering, ``FD" feature disturbance and ``avg. $\Delta$"  average performance improvements.}
    \label{table:performance_different_method}
    \begin{tabular}{lcl|clclclclclclclclcl}
    \hline\noalign{\smallskip}
    \multicolumn{3}{c}{{Module}} & \multicolumn{5}{c}{{Shot}}
    & \multirow{2}{*}{avg. $\Delta$}\\
    \noalign{\smallskip}
    \cline{1-8}
    \noalign{\smallskip}
    MM & FF & FD & \multicolumn{1}{c}1 & \multicolumn{1}{c}2 & \multicolumn{1}{c}3 & \multicolumn{1}{c}5 & 10 & {} \\
    \noalign{\smallskip}
    \hline
    \noalign{\smallskip}
    {} & {} & {} & 14.8 & 15.5 & 26.7 & 33.9 & 47.2 & {} \\ 
    \multicolumn{1}{c}{\checkmark} & {} & {} & 13.5 & 21.9 & 28.5 & 40.2 & 47.0 & \multicolumn{1}{c}{+2.6} \\
    \multicolumn{1}{c}{\checkmark} & {\checkmark} & {} & 13.2 & 23.4 & 29.9 & 43.1 & \bf49.8 & \multicolumn{1}{c}{+4.3}\\
    \multicolumn{1}{c}{\checkmark} & {\checkmark} & \multicolumn{1}{c}{\checkmark} & \bf17.8 & \bf26.1 & \bf31.5 & \bf44.8 & 47.5 & \multicolumn{1}{c}{\bf+5.9}\\
    \hline
    \end{tabular}
    \end{center}
    \end{table}
    \setlength{\tabcolsep}{1.4pt}

\section{Experiments}

\subsection{Experimental Setting}
\textbf{Datasets.}
The proposed CME approach for few-shot object detection is evaluated on Pascal VOC 2007, VOC 2012 and MS COCO,  following the settings in Meta YOLO~\cite{FeatureReweighting}. The object categories in the datasets are divided into two groups: base classes with adequate annotations and novel classes with $K$-shot annotated instances. During base training process, the network is optimized upon using the training data of base classes. During finetuning, the network is optimized by $K$-shot instances of each novel and base classes. For Pascal VOC, the whole dataset is partitioned into 3 splits for cross validation. In each split, 5 classes are selected as novel classes and the rest of the 15 classes are base classes. The number of annotated instances $K$ is set as 1, 2, 3, 5, and 10. For MS COCO, 20 classes are selected as novel ones and the remaining 60 classes are set as base ones. 

\setlength{\tabcolsep}{4pt}
    \begin{table}[t]
    \begin{center}
    \caption{Ablation study of number of output channels in the feature filtering module on Pascal VOC Novel class (split-1).}
    \label{table:Ab_channel_number}
    \begin{tabular}{l|clclclclclclclclclclclclcl}
    \hline\noalign{\smallskip}
    \multicolumn{1}{c}{{\diagbox{Num.}{Shots}}} & \multicolumn{1}{c}{{1}} & \multicolumn{1}{c}{{2}} & \multicolumn{1}{c}{{3}} & \multicolumn{1}{c}{{5}} & \multicolumn{1}{c}{{10}} & \multicolumn{1}{c}{{avg. $\Delta$}}\\
    \noalign{\smallskip}
    \hline
    \noalign{\smallskip}
    {W/O FF}  & {13.5} & {21.9} & {28.5} & {40.2} & {47.0} & {}\\ 
    {1024}  & {13.6} & {19.9} & {27.5} & {36.0} & {48.2} & {-1.2}\\ 
    {512} & {13.2} & \bf{23.5} & \bf{29.9} & \bf{43.1} & \bf{49.8} & \bf{+2.9}\\ 
    {256} & \bf{16.3} & {22.6} & {27.3} & {37.7} & {47.9} & {+1.3}\\ 
    \hline
    \end{tabular}
    \end{center}
    \end{table}
    \setlength{\tabcolsep}{1.4pt}

\setlength{\tabcolsep}{4pt}
    \begin{table}[t]
    \begin{center}
    \caption{Ablation study of self-disturbance on Pascal VOC novel classes (split-1).}
    \label{table:Ab_disturbance_class}
    \begin{tabular}{lc|lclclclclclclclclclcl}
    \hline\noalign{\smallskip}
     \multicolumn{2}{c}{{Method}} & \multicolumn{5}{c}{{Shots}}
    & \multirow{2}{*}{avg. $\Delta$}\\
     \cline{1-7}
    \multicolumn{1}{c}{$C_{Novel}$} & {$C_{Base}$} & \multicolumn{1}{c}{{1}} & \multicolumn{1}{c}{{2}} & \multicolumn{1}{c}{{3}} & \multicolumn{1}{c}{{5}} & \multicolumn{1}{c}{{10}} & {}\\
    \noalign{\smallskip}
    \hline
    \noalign{\smallskip}
    {} & {}  & {13.2} & {23.4} & {29.9} & {43.1} & \bf{49.8} & {}\\ 
    {\checkmark} & {} & {13.0} & {22.7} & {29.5} & {43.5} & {49.5} & {-0.2}\\
    {} & {\checkmark} & \bf{17.8} & \bf{26.1} & \bf{31.5} & \bf{44.8} & {47.5} & \bf{+1.7}\\
    {\checkmark} & {\checkmark} & {16.0} & {24.9} & {31.0} & {43.9} & {49.5} & {+1.2}\\
    \hline
    \end{tabular}
    \end{center}
    \end{table}
    \setlength{\tabcolsep}{1.4pt}

\setlength{\tabcolsep}{4pt}
    \begin{table}[t]
    \begin{center}
    \caption{Comparison of feature disturbance strategies on Pascal VOC novel classes (split-1). ``trun." denotes truncation.} 
    \label{table:Ab_Disturbance_manner}
    \begin{tabular}{l|clclclclclclclclclclclclcl}
    \hline\noalign{\smallskip}
    \multicolumn{1}{c}{{\diagbox{Manner}{Shots}}} & \multicolumn{1}{c}{{1}} & \multicolumn{1}{c}{{2}} & \multicolumn{1}{c}{{3}} & \multicolumn{1}{c}{{5}} & \multicolumn{1}{c}{{10}} & \multicolumn{1}{c}{{avg. $\Delta$}}\\
    \noalign{\smallskip}
    \hline
    \noalign{\smallskip}
    {w/o disturbance}  & {13.2} & {23.4} & {29.9} & {43.1} & \bf{49.8} & {}\\ 
    {Random sample} & {15.2} & {23.2} & {31.4} & {42.2} & {48.8} & {+0.3}\\ 
    {Random crop} & {15.7} & {21.9} & \bf{32.5} & {43.9} & {46.6} & {+0.2}\\ 
    {Feature trun.} & {14.5} & {23.1} & {31.8} & {43.6} & {48.8} & {+0.5}\\ 
    {Gradient trun.} & \bf{17.8} & \bf{26.1} & {31.5} & \bf{44.8} & {47.5} & \bf{+1.7}\\
    \hline
    \end{tabular}
    \end{center}
    \end{table}
    \setlength{\tabcolsep}{1.4pt}

\begin{figure}[t]
\centering
\includegraphics[width=1\linewidth]{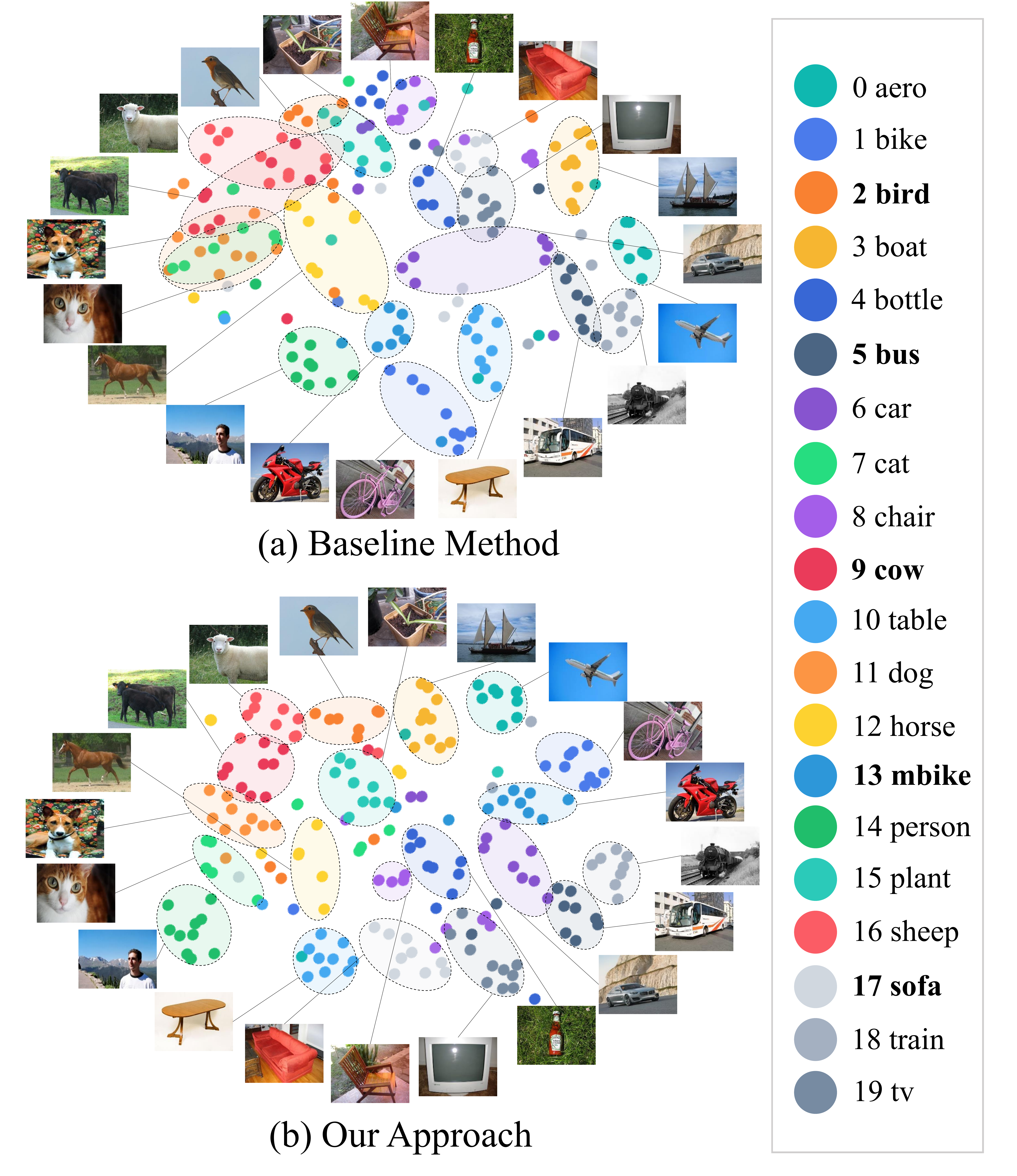}
\caption{t-SNE visualization of prototypes produced by the baseline method~\cite{FeatureReweighting} and our proposed CME approach. The novel classes are in bold font.}
\label{fig:Tsne}
\vspace{-0.2cm}
\end{figure}

\setlength{\tabcolsep}{3pt}
    \begin{table*}[t]
    \begin{center}
    \caption{Detection performance comparison on the Pascal VOC dataset.}
    \label{table:VOC_SOTA}
    \begin{tabular}{l|c|lclcl|clclc|lclclclclclclclclclclcl}
    \hline\noalign{\smallskip}
    \noalign{\smallskip}
    \multicolumn{1}{c}{{}} & \multicolumn{1}{c|}{{}} & \multicolumn{5}{c|}{{Novel set 1}} &  \multicolumn{5}{c|}{{Novel set 2}} & \multicolumn{5}{c}{{Novel set 3}}\\
    \hline\noalign{\smallskip}
     \multirow{1}{*}{Framework} & \diagbox{Method}{Shots} &  \multicolumn{1}{c}{{1}} & \multicolumn{1}{c}{{2}} & \multicolumn{1}{c}{{3}} & \multicolumn{1}{c}{{5}} & \multicolumn{1}{c|}{{10}} & \multicolumn{1}{c}{{1}} & \multicolumn{1}{c}{{2}} & \multicolumn{1}{c}{{3}} & \multicolumn{1}{c}{{5}} & \multicolumn{1}{c|}{{10}} & \multicolumn{1}{c}{{1}} & \multicolumn{1}{c}{{2}} & \multicolumn{1}{c}{{3}} & \multicolumn{1}{c}{{5}} & \multicolumn{1}{c}{{10}}  \\
    \noalign{\smallskip}
    \hline
    \noalign{\smallskip}
    \multirow{4}{*}{YOLO} & {LSTD~\cite{LSTD}} & 8.2 & 11.0 & 12.4 & 29.1 & 38.5 & 11.4 & 3.8 & 5.0 & 15.7 & 31.0 & 12.6 & 8.5 & 15.0 & 27.3 & 36.3\\ 
    & {Meta YOLO~\cite{FeatureReweighting}} & 14.8 & 15.5 & 26.7 & 33.9 & 47.2 & 15.7 & 15.3 & 22.7 & 30.1 & 40.5 & \bf21.3 & 25.6 & 28.4 & 42.8 & 45.9\\ 
    & {MetaDet~\cite{MetaDet}} & 17.1 & 19.1 & 28.9 & 35.0 & \bf48.8 & \bf18.2 & \bf20.6 & 25.9 & 30.6 & \bf41.5 & 20.1 & 22.3 & 27.9 & 41.9 & 42.9\\
    & {\textbf{CME (Ours)}} & \bf17.8 & \bf26.1 & \bf31.5 & \bf44.8 & 47.5 & {12.7} & {17.4} & {\bf27.1} & {\bf33.7} & {40.0} & {15.7} & {\bf27.4} & {\bf30.7} & {\bf44.9} & {\bf48.8}\\
    \hline
    \noalign{\smallskip}
    \multirow{6}{*}{F-RCNN} & {MetaDet~\cite{MetaDet}} & 18.9 & 20.6 & 30.2 & 36.8 & 49.6 & 21.8 & 23.1 & 27.8 & 31.7 & 43.0 & 20.6 & 23.9 & 29.4 & 43.9 & 44.1\\ 
    & {Meta R-CNN~\cite{MetaRCNN}} & 19.9 & 25.5 & 35.0 & 45.7 & 51.5 & 10.4 & 19.4 & 29.6 & 34.8 & 45.4 & 14.3 & 18.2 & 27.5 & 41.2 & 48.1\\ 
    & {Viewpoint~\cite{viewpoint}} & 24.2 & 35.3 & 42.2 & 49.1 & 57.4 & 21.6 & 24.6 & 31.9 & 37.0 & 45.7 & 21.2 & 30.0 & 37.2 & 43.8 & 49.6\\
    & {TFA w/cos~\cite{Frustratingly}} & 39.8 & 36.1 & 44.7 & 55.7 & 56.0 & 23.5 & 26.9 & 34.1 & 35.1 & 39.1 & 30.8 & 34.8 & 42.8 & \textbf{49.5} & 49.8\\
    & {MPSR~\cite{MPSR}} & \textbf{41.7} & {42.5} & \textbf{51.4} & 52.2 & \textbf{61.8} & 24.4 &  29.3 & 39.2 & 39.9 & \textbf{47.8} & \textbf{35.6} & \textbf{41.8} & 42.3 & 48.0 & {49.7}\\
    & {\textbf{CME (Ours)}} & {41.5} & \textbf{47.5} & {50.4} & \textbf{58.2} & {60.9} & \textbf{27.2} & \textbf{30.2} & \textbf{41.4} & \textbf{42.5} & {46.8} & {34.3} & {39.6} & \textbf{45.1} & {48.3} & \textbf{51.5}\\
    \hline
    \end{tabular}
    \end{center}
    \end{table*}
    \setlength{\tabcolsep}{1.4pt}

\textbf{Implementation Details.} 
As a plug-and-play module, CME is fused with the one-stage detector (Meta YOLO~\cite{FeatureReweighting}) and the two-stage detector (MPSR~\cite{MPSR}) for evaluation. In what follows, the experimental analysis and ablation study are based on the Meta YOLO detector, which is implemented with PyTorch 1.0 and run on Nvidia Tesla V100 GPUs. During training, four data augmentation strategies are used, including size normalization, horizontal flipping, random cropping, and random resizing. The network is optimized by the SGD algorithm with initial learning rate of 0.001, momentum of 0.9 for  80,200 iterations in base training and 2000 iterations in finetuning. There are 64 query images per batch and 2 support images for each class.

\subsection{Ablation Study}

Table\ \ref{table:performance_different_method} shows the efficacy of the main components of CME for few-shot object detection with different shot settings on Pascal VOC novel classes (split-1). With the max-margin loss, the average performance gain is $2.6\%$ compared with baseline method. By using feature filtering, the performance gain increases to $4.3\%$. With the feature disturbance for class margin equilibrium, the performance gain increases to $5.4\%$. It shows that CME achieves significant improvement over the baseline method.

\textbf{Max-Margin.} From Table\ \ref{table:performance_different_method}, we can find that max-margin makes effect in 2,3,5 shots setting while being invalid in 1-shot setting. It validates that given limit training data, the increase of class margins is worthless for which is not conducive to the reconstruction of novel class.

\textbf{Feature Filtering.}  In Table~\ref{table:Ab_channel_number}, experiments are conducted to determine the number of output feature channels in the feature filtering module. It reveals that 512 channels reports the best result. 1024 output channels is redundant that makes the margin loss invalid. 256 output channels is insufficiency because of the depression of the feature representation. That is to say that the FC layer requires a slightly smaller output channel number (compared with 1024 input channels) to filter out localization related features.

\textbf{Feature Disturbance.} In Table~\ref{table:Ab_disturbance_class} and Table\ \ref{table:Ab_Disturbance_manner}, ablation studies are carried out to compare the feature disturbance strategies. Table~\ref{table:Ab_disturbance_class} shows that it is better to disturb the prototypes of base class $C_{Base}$ without $C_{Novel}$. According to Eq.\ \ref{eq:equilibrium}, the margin equilibrium is implemented by feature disturbance. The disturbance of base classes depresses margins between base classes which benefit the representation of novel class. Conversely, the disturbance of novel classes may degenerate the representation discrimination as the margin space turns limited. 

Table\ \ref{table:Ab_Disturbance_manner} validates that gradient truncation significantly outperforms feature truncation and feature crop strategies among those of feature disturbance method since it is an adversarial min-max margin manner against the gradient rather than a simple data augmentation strategy.

\begin{figure}[t]
\centering
\includegraphics[width=1\linewidth]{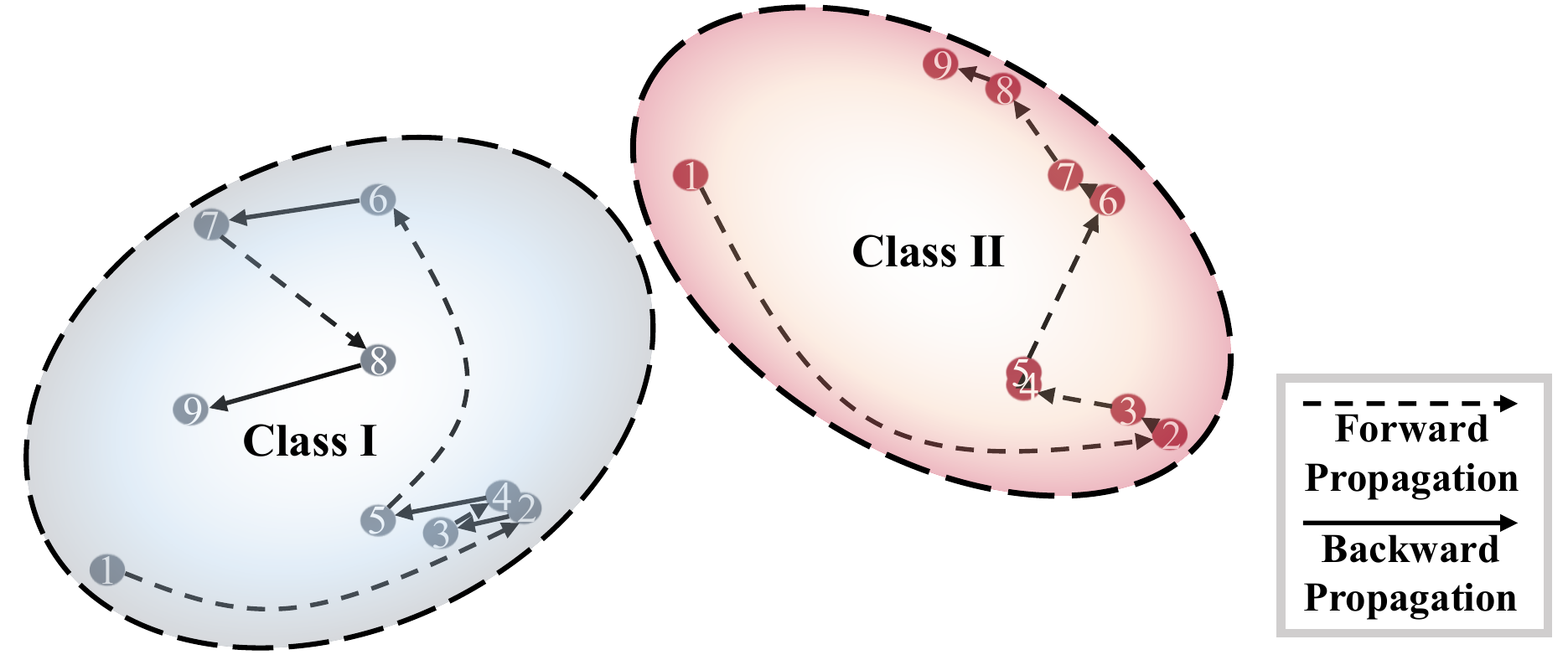}
\caption{t-SNE visualization of the feature prototype evolution of two object classes during the finetuning stage. While the dashed curves denote feature disturbance routes (Forward Propagation), the solid line segments denote the routes driven by the fine-turning loss with max-margin regularization (Backward Propagation). In the finetuning stage, the instance features of novel classes form sub-spaces in the feature space learned on base classes.}
\label{fig:Disturbance}
\end{figure}

\subsection{Model Analysis}
In Fig.\ \ref{fig:Tsne}, we compare the distributions of feature prototypes learned by the baseline method and our CME approach. One can see that CME optimizes novel class embeddings by reserving adequate margin space for novel classes when learning the feature representation. Furthermore, CME optimizes feature space partition by pursuing margin equilibrium in an adversarial min-max fashion when finetuning the network with the novel classes. While the baseline method is confused with the novel class ``Cow" and the base classes ``Cat", ``Dog", ``Sheep", and ``Horse", CME clearly distinguishes them and reduces the overlap between classes. It proves that the CME can improve the representation capacity of the feature space for better object detection. 

Fig.\ \ref{fig:Disturbance} visualizes the evolution of two prototypes in the feature space during the finetuning stage. With an adversarial min-max margin way (defined as Eq.\ \ref{eq:equilibrium}), during forward propagation, the feature prototypes disturb to minimize the margin which is implemented by re-sampling the support mask. During backward propagation, the feature prototypes move to maximum the margin which is driven by the finetuning loss (Eq.\ \ref{eq:loss_finetune}). With multiple forward-backward propagation iterations, the samples span a feature sub-space for each object class.

\begin{figure}[t]
\centering
\includegraphics[width=1\linewidth]{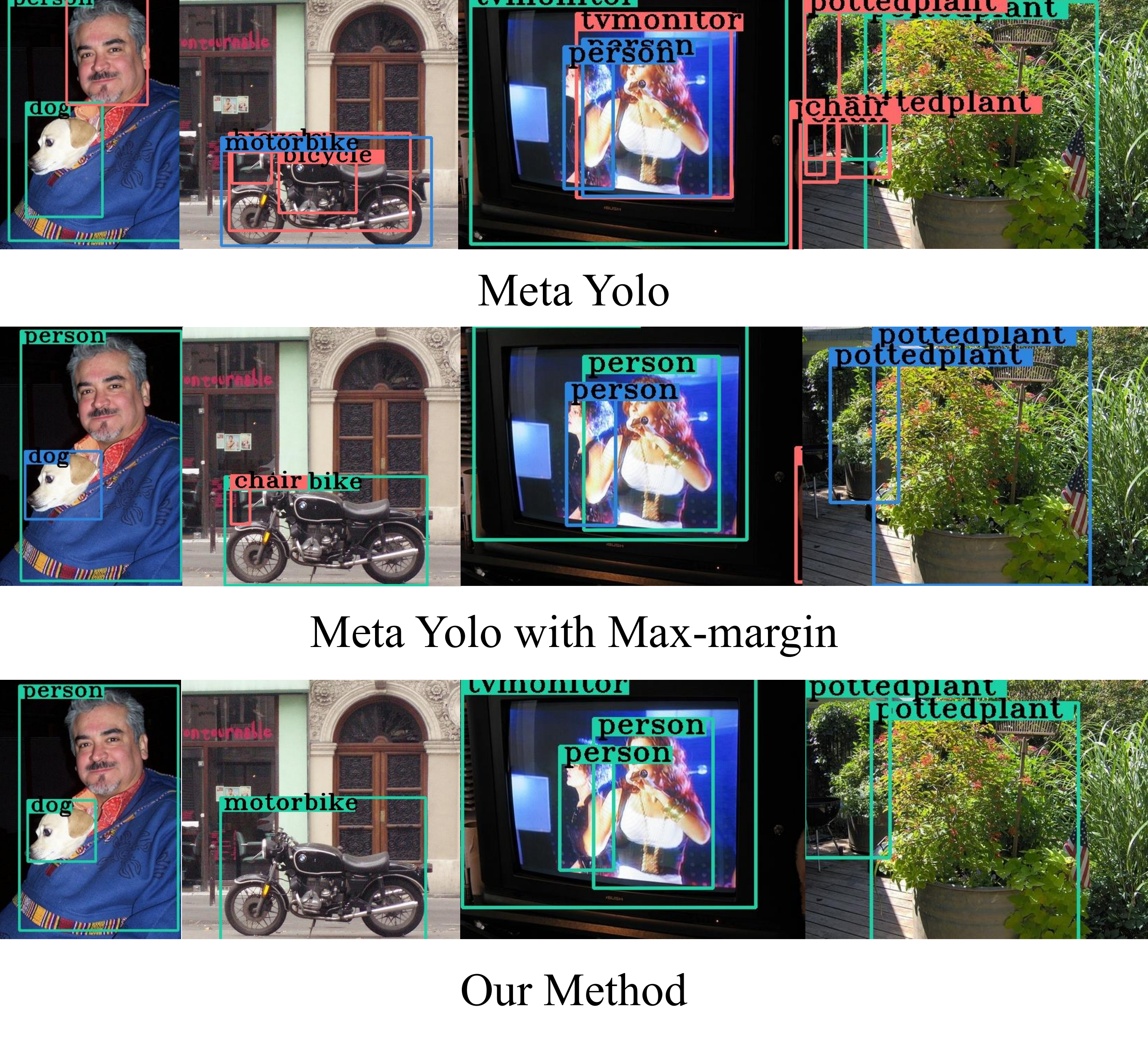}
\caption{Comparison of detection results of the baseline method and the proposed CME approach. Red boxes indicate false detection results, green  boxes indicate true detection results and blue boxes indicate missed objects.}
\label{fig:examples}
\end{figure}

Fig.\ \ref{fig:examples} shows the detection result of the baseline method and the proposed CME approach. With the max-margin loss, the few-shot detector can reduce the false detection results because the margin between each class is increased which benefit the discrimination of the classifier. However, the max-margin is not conductive to the feature reconstruction with the raise of missing object. By margin-equilibrium, our approach balanced the contradiction between classification and representation. It shows that CME can precisely detect more objects with fewer false positives.

\setlength{\tabcolsep}{3pt}
    \begin{table*}[t]
    \begin{center}
    \caption{Performance comparison on the MS COCO dataset.}
    \label{table:COCO_SOTA}
    \begin{tabular}{l|l|ccccccccccccccccccccccccccc}
    \hline\noalign{\smallskip}
 \multicolumn{1}{c|}{Shots} & \makecell[c]{Method} &  \multicolumn{1}{c}{{AP}} & \multicolumn{1}{c}{{AP50}} & \multicolumn{1}{c}{{AP75}} & \multicolumn{1}{c}{{APS}} & \multicolumn{1}{c}{{APM}} & \multicolumn{1}{c}{{APL}} & \multicolumn{1}{c}{{AR1}} & \multicolumn{1}{c}{{AR10}} & \multicolumn{1}{c}{{AR100}} & \multicolumn{1}{c}{{ARS}} & \multicolumn{1}{c}{{ARM}} & \multicolumn{1}{c}{{ARL}} &   \\
    \noalign{\smallskip}
    \hline
    \noalign{\smallskip}
    \multirow{7}{*}{10} & {LSTD~\cite{LSTD}} & 3.2 & 8.1 & 2.1 & 0.9 & 2.0 & 6.5 & 7.8 & 10.4 & 10.4 & 1.1 & 5.6 & 19.6\\ 
    & {Meta YOLO~\cite{FeatureReweighting}} & 5.6 & 12.3 & 4.6 & 0.9 & 3.5 & 10.5 & 10.1 & 14.3 & 14.4 & 1.5 & 8.4 & 28.2 \\ 
    & {MetaDet~\cite{MetaDet}} & 7.1 & 14.6 & 6.1 & 1.0 & 4.1 & 12.2 & 11.9 & 15.1 & 15.5 & 1.7 & 9.7 & 30.1\\
    & {Meta R-CNN~\cite{MetaRCNN}} & 8.7 & 19.1 & 6.6 & 2.3 & 7.7 & 14.0 & 12.6 & 17.8 & 17.9 & \bf7.8 & 15.6 & 27.2 \\
    & {TFA w/cos~\cite{Frustratingly}} & 10.0 & - & 9.3 & - & - & - & - & - & - & - & - & - \\
    & {Viewpoint~\cite{viewpoint}} & 12.5 & \bf27.3 & 9.8 & 2.5 & 13.8 & 19.9 & \bf20.0 & \bf25.5 & \bf25.7 & 7.5 & \bf27.6 &38.9 \\
    & {MPSR~\cite{MPSR}} & 9.8 & 17.9 & 9.7 & 3.3 & 9.2 & 16.1 & 15.7 & 21.2 & 21.2 & 4.6 & 19.6 & 34.3 \\
    & {\textbf{CME (Ours)}} & \bf15.1 & 24.6 & \bf16.4 & \bf4.6 & \bf16.6 & \bf26.0 & 16.3 & 22.6 & 22.8 & 6.6 & 24.7 & \bf39.7 \\
    \hline
    \noalign{\smallskip}
    \multirow{7}{*}{30} & {LSTD~\cite{LSTD}} & 6.7 & 15.8 & 5.1	& 0.4 & 2.9	& 12.3 & 10.9 & 14.3 & 14.3	& 0.9 &7.1 & 27.0 \\ 
    & {Meta YOLO~\cite{FeatureReweighting}} & 9.1	& 19.0 & 7.6 & 0.8 & 4.9 & 16.8 & 13.2 & 17.7 & 17.8 & 1.5 & 10.4 & 33.5 \\ 
    & {MetaDet~\cite{MetaDet}} & 11.3 & 21.7 & 8.1 & 1.1 & 6.2 & 17.3 & 14.5 & 18.9 & 19.2 & 1.8 & 11.1 & 34.4 \\
    & {Meta R-CNN~\cite{MetaRCNN}} & 12.4 & 25.3 & 10.8 & 2.8 & 11.6 & 19.0 & 15.0 & 21.4 & 21.7 & \bf8.6 & 20.0 & 32.1 \\
    & {TFA w/cos~\cite{Frustratingly}} & 13.7 & - & 13.4 & - & - & - & - & - & - & - & - & - \\
    & {Viewpoint~\cite{viewpoint}} & 14.7 & \bf30.6 & 12.2 & 3.2 & 15.2 & 23.8 & \bf22.0 & \bf28.2 & \bf28.4 & 8.3 & \bf30.3 & 42.1\\
    & {MPSR~\cite{MPSR}} & 14.1 & 25.4 & 14.2 & 4.0 & 12.9 & 23.0 & 17.7 & 24.2 & 24.3 & 5.5 & 21.0 & 39.3\\
    & {\textbf{CME (Ours)}} & \bf16.9 & 28.0 & \bf17.8 & \bf4.6 & \bf18.0 & \bf29.2 & 17.5 &23.8 &24.0 &6.0 &24.6 & \bf42.5
 \\
    \hline
    \end{tabular}
    \end{center}
    \end{table*}
    \setlength{\tabcolsep}{1.4pt}

\subsection{Performance Comparison}

\textbf{Pascal VOC}
In Table~\ref{table:VOC_SOTA}, we compare CME with the one-stage few-shot detectors including LSTD~\cite{LSTD}, Meta YOLO~\cite{FeatureReweighting}, and MetaDet~\cite{MetaDet}, which are based on the YOLO detector. 
The proposed CME detector demonstrates great advantages over the compared detectors.
Specifically, for Novel Set 1, CME respectively achieves 0.7\%(17.8\% vs. 17.1\%) on 1-shot setting, 7.0\%(26.1\% vs. 19.1\%) on 2-shot setting,  2.6\%(31.5\% vs. 28.9\%) on 3-shot setting, 9.8\%(44.8\% vs. 35.0\%) on 5-shot setting. The average improvement is 3.8\%, which is s significant margin for the challenging task. The average performance improvements are respectively {0.7\%} for novel set 3.

We also compare the proposed approach with two-stage detectors including  MetaDet~\cite{MetaDet}, Meta RCNN~\cite{MetaRCNN}, TFA~\cite{Frustratingly} Viewpoint Estimation~\cite{viewpoint}, and MPSR~\cite{MPSR}, which are based on the Faster-RCNN framework. One can see that in most settings CME outperforms the compared detectors. 
For novel set 1, CME respectively outperforms by 5\%(47.5\% vs. 42.5\%) on 2-shot setting,  6\%(58.2\% vs. 52.2\%) on 5-shot setting. The average improvement researches 1.2\%. The average performance improvement for novel set 2 is 1.5\% and 0.3\% for novel set 3.

\textbf{MS COCO}
Compared with Pascal Voc, MS COCO has more object categories and images, which imply that the margin equilibrium may benefit for much richer feature representation. Thereby,  our approach achieves more significant relative improvement on MS COCO as shown in Table~\ref{table:COCO_SOTA}. For the 10-shot setting, CME improves  AP upon the baseline method MPSR by 5.3\% and for the 30-shot setting, it improves 2.8\%. 

\section{Conclusion}

We proposed a class margin equilibrium (CME) approach to optimize both feature space partition and novel class representation for few-shot object detection. During base training, CME preserves adequate margin space for novel classes by a simple-yet-effective class margin loss. During finetuning, CME pursues margin equilibrium by disturbing the instance features of novel classes in an adversarial min-max fashion. Extensive experiments validated the effectiveness of CME for alleviating the constriction of feature representation and classification in few-shot settings. As a plug-and-play module, CME improved both one-stage and two-stage few-shot detectors, in striking contrast to the state-of-the-arts. As a general method for feature representation learning and class margin optimization, CME provides a fresh insight for few-shot learning problems.

\textbf{Acknowledgement.} This work was supported by Natural Science Foundation of China (NSFC) under Grant 61836012 and 61771447, the Strategic Priority ResearchProgram of Chinese Academy of Sciences under Grant No. XDA27000000, CAAI-Huawei MindSpore Open Fund and MindSpore deep learning computing framework at \href{https://www.mindspore.cn} {\color{magenta}https://www.mindspore.cn}

{\small
\bibliographystyle{ieee_fullname}
\bibliography{egbib}
}

\end{document}